\documentclass{article}

\usepackage{arxiv}

\usepackage[utf8]{inputenc} % allow utf-8 input
\usepackage[T1]{fontenc}    % use 8-bit T1 fonts
\usepackage{hyperref}       % hyperlinks
\usepackage{url}            % simple URL typesetting
\usepackage{booktabs}       % professional-quality tables
\usepackage{amsfonts}       % blackboard math symbols
\usepackage{nicefrac}       % compact symbols for 1/2, etc.
\usepackage{microtype}      % microtypography
\usepackage{lipsum}
\usepackage{graphicx}
\usepackage{amsmath}
\graphicspath{ {./images/} }

\usepackage{todonotes}

\title{LLaVA-Zip: Adaptive Visual Token Compression with Intrinsic Image Information}

\author{
 Ke Wang \\
  University of Notre Dame \\
  South Bend, IN 46556 \\
  \texttt{kwang5@nd.edu} \\
   \And
 Hong Xuan \\
  TikTok\\
  Bellevue, WA 98004 \\
  \texttt{hong.xuan@bytedance.com} \\
 \\
}

\begin{document}
\maketitle
\begin{abstract}
Multi-modal large language models (MLLMs) utilizing instruction-following data, such as LLaVA, have achieved great progress in the industry. A major limitation in these models is that visual tokens consume a substantial portion of the maximum token limit in large language models (LLMs), leading to increased computational demands and decreased performance when prompts include multiple images or video. Industry solutions often mitigate this issue by increasing computational power, but this approach is less feasible in academic environments with limited resources. In this study, we propose Dynamic Feature Map Reduction (DFMR) based on LLaVA-1.5 to address the challenge of visual token overload. DFMR dynamically compresses the visual tokens, freeing up token capacity. Our experimental results demonstrate that integrating DFMR into LLaVA-1.5 significantly improves the performance of LLaVA in varied visual token lengths, offering a promising solution for extending LLaVA to handle multi-image and video scenarios in resource-constrained academic environments and it can also be applied in industry settings for data augmentation to help mitigate the scarcity of open-domain image-text pair datasets in the continued pretraining stage.
\end{abstract}

% keywords can be removed
% \keywords{MLLM \and Second keyword \and More}

\begin{figure}[h]
    \centering
    \includegraphics[width=1.0\linewidth]{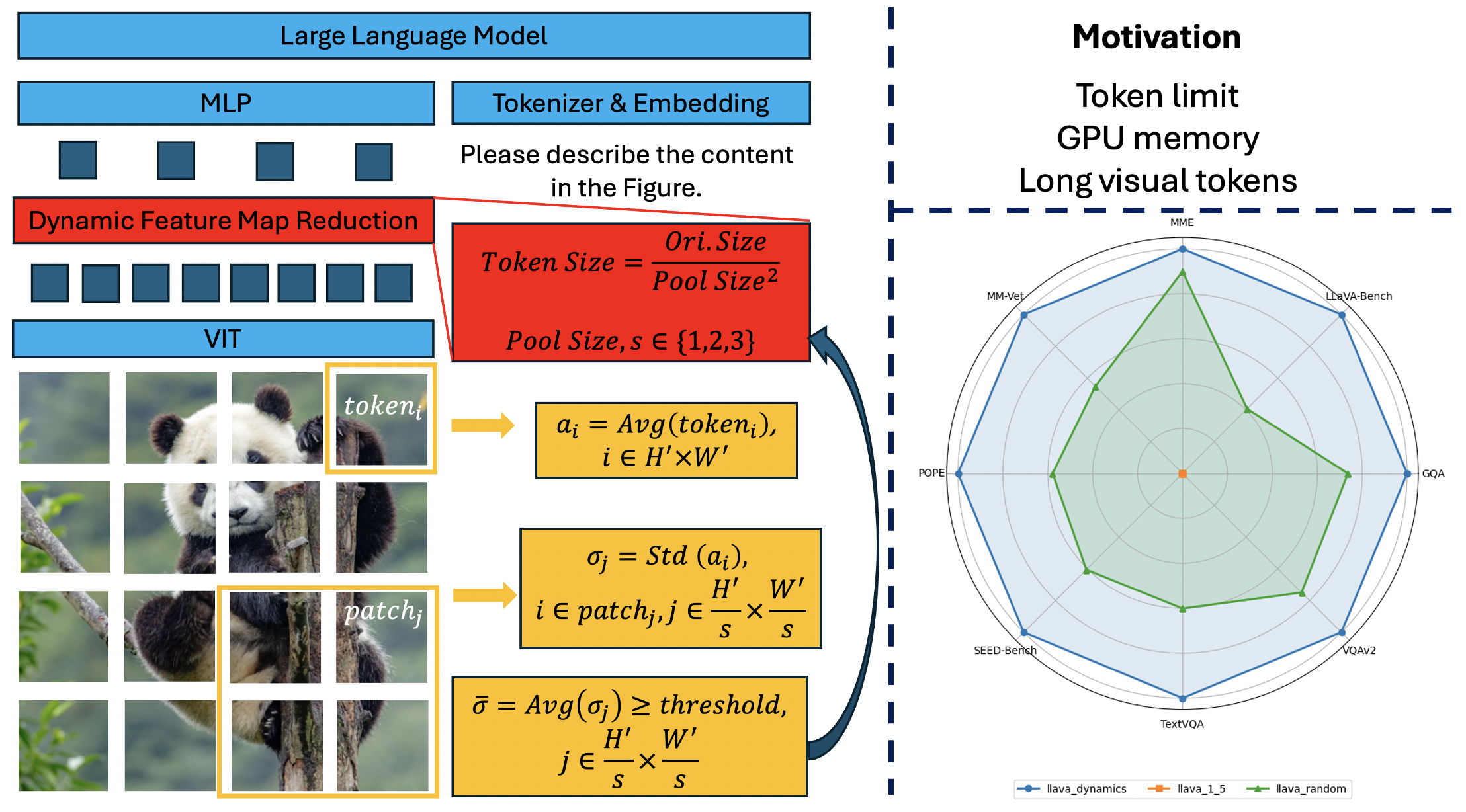}
    \caption{\textbf{Overall Structure of LLaVA-Dynamics.} The core module is the Dynamic Feature Map Reduction (DFMR), which dynamically compresses the visual tokens from the visual encoder based on the calculated image information. \textbf{Results Summary.} The proposed DFMR module demonstrates clear improvements in all benchmark tasks over the baseline and the original LLaVA-1.5, see Table \ref{tab:res_summary}.}
    \label{fig:model_summary}
\end{figure}

\section{Introduction}

Recently, the multimodal large language models (MLLMs), especially based on instruction following data, LLaVA family\cite{li2024llava,liu2024improved,liu2024visual}, have achieved great progress in the industry. One of the major limitations of these models is that the visual tokens consume a substantial portion of the maximum token limit in the pre-trained large language model (LLM). To address this, researchers apply the AnyRes method\cite{li2024llava}, which splits the single input image into multiple sub-images and uses the visual tokenizer to tokenize all the sub-images and optionally the original images. The method has proved effective and can be widely applied in industrial settings with sufficient computation power. However, in academic settings, where computing resource is constrained. The increase of visual tokens is limited by the maximum GPU memory and maximum token length in the pre-trained LLM.

To address the problems, researchers have concentrated on reducing the number of visual tokens. Li et al.\cite{li2023blip} introduced BLIP-2, which utilizes a Q-former to compress image information into a fixed set of visual tokens, effectively condensing the visual input. Recent works have proposed methods that leverage textual guidance to identify and retain important visual tokens\cite{song2024less,zhang2024beyond}. Arif et al.\cite{arif2024hired} presented HiRED, an attention-guided token dropping technique that employs CLIP attention masks to split images and selectively discard less relevant tokens, enhancing efficiency in resource-constrained environments. Additionally, Cai et al.\cite{cai2024matryoshka} developed Matryoshka Multimodal Models, which manage image representations by organizing visual tokens in a coarse-to-fine hierarchy. Generally, the current approach focuses on the use of external information outside of the image itself to facilitate the compression of visual tokens, instead of fully leveraging the information the image can provide.

In this paper, we propose DFMR (Dynamic Feature Map Reduction), a method that dynamically adjusts the compression ratio of visual tokens based solely on the intrinsic content of each image. By introducing a novel metric that quantifies the information within an image, DFMR determines the optimal number of visual tokens required for effective representation. This enables the model to allocate computational resources more efficiently—focusing on complex images rich in information while compressing simpler ones to reduce redundancy.

To validate the effectiveness of DFMR, we conduct experiments on various mainstream multimodal evaluation benchmarks. The results demonstrate that our method improves performance across all eight tasks compared with LLaVA-1.5 and our designed baseline in compressed visual tokens. In summary, the key contributions of this paper are:

\begin{itemize}
    \item We propose DFMR for dynamically compressing visual tokens based on the intrinsic information of images.

    \item We develop a metric to quantify image information, guiding compression ratio determination.

    \item We show that DFMR improves model performance in scenarios involving compressed visual tokens.
\end{itemize}

\section{Related Work}

\textbf{MLLMs.}
The most promising approach to bridging visual and textual information involves integrating visual information into LLMs. Flamingo\cite{alayrac2022flamingo} proposed using cross-attention mechanisms to incorporate encoded image information directly into LLM layers. In contrast, BLIP-2\cite{li2023blip} introduced a Q-Former to effectively bridge encoded visual information into the input space of LLMs. LLaVA\cite{li2024llava,liu2024improved,liu2024visual} adopted a simpler approach by utilizing a multilayer perceptron (MLP) to map encoded visual information into the LLM's input space while incorporating high-quality data and adaptive image segmentation techniques to handle multiple images and videos efficiently.

\textbf{Visual token compression.} The compression of visual tokens can be broadly categorized into three approaches: QFormer-like, LLM-assisted, and handcrafted metrics methods \cite{zhu2024focusllava}. The QFormer-like method \cite{li2023blip,bai2023qwen} maps any number of images to a fixed number of visual tokens, which often results in information loss for tasks requiring fine-grained details. The LLM-assisted method \cite{song2024less,zhang2024beyond} utilizes LLMs to perform visual token pruning but does not fully exploit the internal information pre-trained from images. The heuristic approach \cite{arif2024hired,cai2024matryoshka} relies on handcrafted metrics to evaluate the importance of visual tokens. Our proposed method falls into the third category, where we propose a metric to measure the intrinsic information represented by the image and dynamically compress the image based on this metric.

\section{Method}

\subsection{Overall Structure}

The overall structure of the proposed model is illustrated in Fig. \ref{fig:model_summary}. Our model is developed based on LLaVA-1.5\cite{liu2024improved}, with an additional module, Dynamic Feature Map Reduction (DFMR), inserted between the vision encoder and the projector. The DFMR dynamically compresses the visual tokens based on the image's inherent information.

Given a resized input given image \( \mathbf{I} \in \mathbb{R}^{H \times W \times 3} \), it is passed through the vision encoder to obtain visual tokens \( \mathbf{V} \in \mathbb{R}^{N_v \times D_v } \), where \( N_v = H' \times W' \) is the number of visual tokens determined by the image feature map size \( H' \) and \( W' \) after the vision encoder, and \( D_v \) is the embedding dimension of the visual tokens. The obtained visual tokens \( \mathbf{V} \) are then passed through the proposed DFMR module to produce compressed visual tokens \( \mathbf{V}' \in \mathbb{R}^{N_c \times D_v} \), where \( N_c = \left( \frac{H'}{s} \times \frac{W'}{s} \right) \) and \( s \) is the compression factor of dynamic pooling. The compressed visual tokens \( \mathbf{V}' \) are flattened and connected to the LLM via the projector, resulting in projected visual tokens \( \mathbf{V}^p \in \mathbb{R}^{N_c \times D_l} \), where \( D_l \) is the embedding dimension of the LLM.

For a given text input, we tokenize it to obtain textual tokens \( \mathbf{T} \in \mathbb{R}^{N_t \times D_l} \), where \( N_t \) is the number of textual tokens. The projected visual tokens \( \mathbf{V}^p \) and textual tokens \( \mathbf{T} \) are concatenated to form the input sequence to the LLM \( \mathbf{S} = [\mathbf{V}^p; \mathbf{T}] \in \mathbb{R}^{(N_c + N_t) \times D_l} \), which is then fed into the LLM for next token prediction.

\subsection{Dynamic Feature Map Reduction (DFMR)}

DFMR conducts an $s \times s$ average pooling operation on the original visual tokens \( \mathbf{V} \in \mathbb{R}^{N_v \times D_v} \). 

First, it divides \( \mathbf{V} \) into non-overlapping windows, each window size is \( p \times p \), and \( p = \frac{H'}{s} = \frac{W'}{s} \) is the window length. Inside each window \( \mathbf{W}_k \), where \( k \) indexes the windows, we calculate the standard deviation \( \sigma_k \) of the values within that patch:

\begin{equation}
\sigma_k = \sqrt{\frac{1}{p^2} \sum_{(u,v) \in \mathbf{W}_k} \left( V_{u,v} - \mathbf{\mu}_k \right)^2},    
\end{equation}

where \( V_{u,v} \) is the feature at position \( (u, v) \) within patch \( \mathbf{W}_k \), and \( \mathbf{\mathbf{\mu}_k} \) is the mean of the patch, given by:

\begin{equation}
\mathbf{\mu}_k = \frac{1}{n_p} \sum_{(u,v) \in \mathbf{W}_k} V_{u,v}.    
\end{equation}

After computing the standard deviation for all patches, we calculate the mean of these patches standard deviations:

\begin{equation}
\overline{\sigma} = \frac{1}{K} \sum_{k=1}^{K} \sigma_k,    
\end{equation}

where \( K \) is the total number of patches in \( \mathbf{V} \).

Based on the value of \( \overline{\sigma} \), the DFMR dynamically determines the appropriate compression factor \( s \) to average pool the original visual tokens $\mathbf{V}$. A higher \( \overline{\sigma} \) indicates greater variability within the image patches, suggesting that less compression (smaller pooling size) is needed to preserve important details. Conversely, a lower \( \overline{\sigma} \) allows for more aggressive compression (larger pooling size) without significant loss of information. A manually defined threshold \( \tau \) determines when to stop increasing the compression factor $s$. For each original visual token \( \mathbf{V} \), the process starts with the smallest compression factor and incrementally increases it until \( \overline{\sigma} \) exceeds \( \tau \) or $s$ has reached the pre-defined maximum value. Once the process stops, the current $s$ is chosen as the final compression factor.

After selecting the appropriate compression factor \(s\), the average pooling operation is applied to the original feature map 
\(\mathbf{V}\). Each element in the resulting compressed feature map 
\(\mathbf{V}' \in \mathbb{R}^{(H'/s) \times (W'/s) \times D_v}\) is given by the average of the corresponding 
\(s \times s\) window in \(\mathbf{V}\):

\[
V'_{i,j,c} = \frac{1}{s^2} \sum_{u=0}^{s-1} \sum_{v=0}^{s-1} V_{s \cdot i + u, \, s \cdot j + v, \, c},
\]

where \(c \in \{0, 1, \ldots, D_v - 1\}\) is the index of embedding, and 
\(i \in \{0, 1, \ldots, (H'/s)-1\}\), \(j \in \{0, 1, \ldots, (W'/s)-1\}\) index the spatial positions in the pooled output.

\section{Experiments}
\subsection{Implementation Details}

We use CLIP ViT-L/14\cite{radford2021learning} as the visual encoder (default resolution 336 $\times$ 336), Vicuna-7B\cite{zheng2023judging} as the LLM, and a 2-layer MLP as the connector. Images are resized to 336 $\times$ 336 before being fed into CLIP. We set the DFMR between CLIP and the MLP with a fixed threshold of 5e-2, where the compression ratio in DFMR for each image is dynamically determined by the threshold and mean of standard deviation $\bar{\sigma}$. Following LLaVA-1.5\cite{liu2024improved}, we use the same training settings for pre-training and visual instruction fine-tuning. In both stages, a cosine scheduler is applied, and the learning rate is set to 1e-3 with a batch size of 256 during pre-training, and 2e-5 with a batch size of 128 during fine-tuning. Training is conducted for one epoch using the AdamW optimizer. Evaluation is conducted using the lmms-eval\cite{lmms_eval2024} framework with a batch size of 1. DeepSpeed ZeRO3 is utilized as the training framework. We conduct experiments on a server equipped with 8 NVIDIA H100 GPUs, each with 80 GB of VRAM.

\textbf{Training Details.} We train three types of models for comparison: LLaVA-1.5, LLaVA-1.5 Random, and LLaVA-1.5 DFMR. For LLaVA-1.5, the number of visual tokens remains uncompressed and is fixed at $576$ for all images during training. For LLaVA-1.5 Random, the number of visual tokens is randomly compressed for each image during training, with the compression factor \( s \) uniformly sampled from the set \( \{1, 2, 3\} \). This results in the number of visual tokens being $576$, or reduced to $144$ and $64$, respectively. For LLaVA-1.5 DFMR, the compression of visual tokens is determined based on our proposed DFMR method for each image during training and it also achieves compression to $576$, $144$, and $64$ visual tokens, respectively.

\textbf{Evaluation Details.} For each model, we evaluate its performance on images with compression ratios of $1$, $2$, and $3$, corresponding to $576$, $144$, and $64$ token scenarios, respectively. These settings effectively evaluate model performance under constrained GPU memory or when the prompt reaches the maximum token length of an LLM. For instance, when the original number of visual tokens is $576$, limited GPU memory or the maximum token length may require compressing the $576$ tokens into $144$ or $64$ tokens.

\subsection{Datasets and Benchmark}

\textbf{Training datasets.} We pre-train and fine-tune the proposed model following LLaVA-1.5, using llava-pretrain-558k for the pre-training stage and llava-instruct for the fine-tuning stage.

\textbf{Benchmarks.} We use 8 popular benchmarks to evaluate our method, including (1) General question-answering benchmarks such as GQA\cite{hudson2019gqa}; (2) Optical character-based visual question-answering benchmarks such as TextVQA\cite{singh2019towards} and VQAv2\cite{goyal2017making}; (3) MLLM benchmarks for specific abilities, like POPE\cite{li2023evaluating}, MM-Vet\cite{yu2023mm}, and LLaVA-Bench\cite{liu2024improved}; (4) Comprehensive MLLM benchmarks such as MME\cite{Fu2023MMEAC} and SEED-Bench\cite{li2023seed}.

\subsection{Performance}

Table \ref{tab:res_summary} summarizes our experimental results. When comparing LLaVA-1.5 random with LLaVA-1.5, we observe that incorporating various visual token lengths enhances the performance of LLaVA-1.5 across different visual token inputs. Since the prior model is regularly exposed to various compressed visual tokens during training, we designate it as our baseline. Comparing LLaVA-1.5 DFMR with LLaVA-1.5, we find that the performance of LLaVA-1.5 DFMR using the original 576 visual tokens is slightly lower than that of the original LLaVA-1.5. This outcome is reasonable given that LLaVA-1.5 DFMR is exposed to a smaller number of original visual tokens. However, when comparing LLaVA-1.5 DFMR with LLaVA-1.5 random, we observe that our proposed DFMR enhances model performance across all tasks and various visual token scenarios.

On average, we achieve significantly better results compared to both LLaVA-1.5 and LLaVA-1.5 random. Our findings indicate that our implemented baseline is highly competitive relative to LLaVA-1.5, and our proposed DFMR demonstrates clear improvements over this baseline, confirming the effectiveness of the proposed DFMR method.

% \begin{table}[!h]
%  \caption{Comparison of models on popular benchmarks. GQA\cite{hudson2019gqa}, LLaVA-Bench\cite{liu2024improved}, MME\cite{Fu2023MMEAC}, MM-Vet\cite{yu2023mm}, POPE\cite{li2023evaluating}, SEED-Bench\cite{li2023seed}, TextVQA\cite{singh2019towards}, and VQAv2\cite{goyal2017making}}
%   \centering
%   \begin{tabular}{lllllllll}
%     \toprule
%     Method& GQA& LLaVA-Bench& MME& MM-Vet& POPE& SEED-Bench& TextVQA& VQAv2\\
%     \midrule
%     LLaVA-1.5& 57.32& 54.63& 1319.85& 25.96& 82.31& 59.19& 30.09& 69.65\\
%     LLaVA-1.5 Random& 60.23& 57.9& 1420.13& 28.43& 84.19& 62.08& 35.66& 73.57\\
%     LLaVA-1.5 DFMR& \textbf{61.27}& \textbf{62.67}& \textbf{1431.38}& \textbf{30.46}& \textbf{85.56}& \textbf{63.96}& \textbf{39.36}& \textbf{74.88}\\
%   \end{tabular}
%   \label{tab:res_summary}
% \end{table}

\begin{table}[ht]
\caption{Comparison of models on popular benchmarks. GQA\cite{hudson2019gqa}, LLaVA-Bench\cite{liu2024improved}, MME\cite{Fu2023MMEAC}, MM-Vet\cite{yu2023mm}, POPE\cite{li2023evaluating}, SEED-Bench\cite{li2023seed}, TextVQA\cite{singh2019towards}, and VQAv2\cite{goyal2017making}. The first column indicates the number of visual tokens for each image when setting the evaluation.}
\centering
\begin{tabular}{lccccccccc}
\toprule
\# of Tokens  & GQA & LLaVA-Bench & MME & MM-Vet & POPE & SEED-Bench & TextVQA & VQAv2 \\
\midrule
\multicolumn{9}{c}{\textbf{LLaVA-1.5}} \\
\midrule
576 & 62.78 & 62.10 & 1490.52 & 33.62 & 86.33 & 67.26 & 46.82 & 77.23 \\
144 & 56.63 & 54.30 & 1358.08 & 24.77 & 82.47 & 58.71 & 27.08 & 69.54 \\
64  & 52.55 & 47.50 & 1110.94 & 19.49 & 78.14 & 51.61 & 16.38 & 62.17 \\
Average & 57.32 & 54.63 & 1319.85 & 25.96 & 82.31 & 59.19 & 30.09 & 69.65 \\
\midrule
\multicolumn{9}{c}{\textbf{LLaVA-1.5 Random}} \\
\midrule
576 & 60.07 & 58.40 & 1410.97 & 29.36 & 84.33 & 62.22 & 35.67 & 73.55 \\
144 & 60.20 & 57.60 & 1443.53 & 28.03 & 84.42 & 62.10 & 35.62 & 73.53 \\
64  & 59.89 & 57.70 & 1405.89 & 27.89 & 83.31 & 61.91 & 35.68 & 73.62 \\
Average & 60.23 & 57.90 & 1420.13 & 28.43 & 84.19 & 62.08 & 35.66 & 73.57 \\
\midrule
\multicolumn{9}{c}{\textbf{LLaVA-1.5 DFMR}} \\
\midrule
576 & 62.66 & 63.90 & 1448.64 & 32.66 & 86.47 & 66.63 & 47.13 & 76.89 \\
144 & 61.42 & 61.80 & 1455.50 & 29.95 & 85.11 & 63.95 & 38.91 & 75.00 \\
64  & 59.72 & 62.30 & 1390.00 & 28.76 & 85.10 & 61.30 & 32.03 & 72.74 \\
Average & \textbf{61.27} & \textbf{62.67} & \textbf{1431.38} & \textbf{30.46} & \textbf{85.56} & \textbf{63.96} & \textbf{39.36} & \textbf{74.88} \\
\bottomrule
\end{tabular}
\label{tab:res_summary}
\end{table}

\begin{figure}[h]
    \centering
    \includegraphics[width=1.0\linewidth]{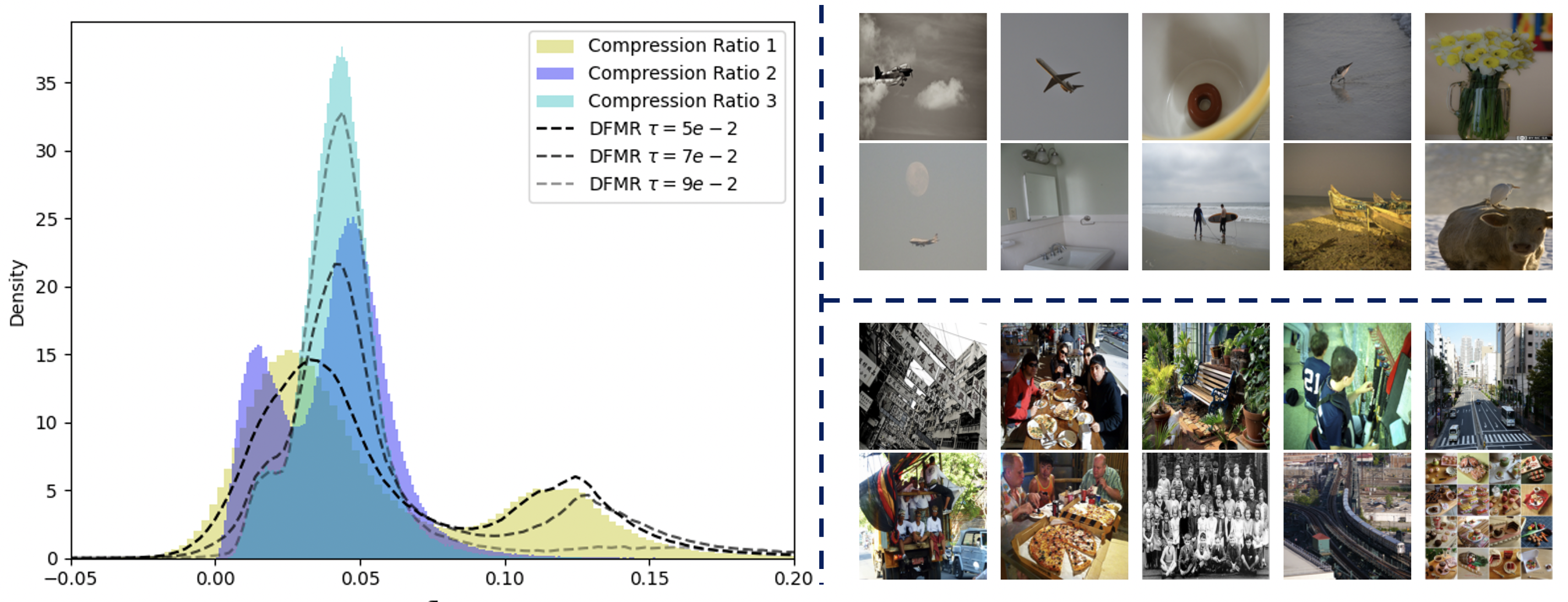}
    \caption{\textbf{Data Distribution.} The figure illustrates the distribution of the calculated standard deviation metric under different compression ratios $s \in \{1, 2, 3\}$ and using the proposed DFMR with manually defined threshold $\tau \in \{5e^{-2}, 7e^{-2}, 9e^{-2}\}$. \textbf{Bottom 10 Samples.} These samples correspond to the 10 images with the smallest calculated mean standard deviation metric $\bar{\sigma}$, typically consisting of repetitive or uniform content and thus more amenable to compression without significant loss of information. \textbf{Top 10 Samples.} These samples correspond to the 10 images with the largest calculated mean standard deviation metric $\bar{\sigma}$, representing those with rich, fine-grained details requiring higher fidelity during compression.}
    \label{fig:model_analysis}
\end{figure}

\subsection{Analysis}
We compute the standard deviations $\sigma$ across patches for each image in the COCO validation dataset \cite{lin2014microsoft} for the analysis experiments conducted in this section. 

Figure \ref{fig:model_analysis}(a) presents the distribution of the calculated results for all images. When the compression ratio is set to 1 or 2, the resulting distributions resemble a mixture of Gaussian distributions with two distinct centers. In contrast, the compression ratio of 3 produces more concentrated distributions that approximate a single Gaussian. This observation suggests that the intrinsic information content varies across images and that compressing visual tokens using a fixed compression ratio affects images unevenly. While larger compression ratios efficiently compress images characterized by repetitive token patterns, they risk losing important fine-grained details in more complex images.

In contrast, our Dynamic Feature Mixing Ratio (DFMR) method dynamically adapts the compression of visual tokens based on each image’s intrinsic information. The black dashed lines in Figure \ref{fig:model_analysis}(a) demonstrate that DFMR effectively compresses images containing redundant information while preserving subtle details in more intricate images. The overall data compression ratio can be adjusted by manually setting the threshold value $\tau$ and can also be configured proportionally to the learning rate for data augmentation training during the continued pretraining stage.

Figures \ref{fig:model_analysis}(b) and \ref{fig:model_analysis}(c) further illustrate this point by displaying the top 10 and bottom 10 images ranked by their calculated standard deviations, respectively. In Figure \ref{fig:model_analysis}(b), many images are predominantly composed of background content or repetitive patterns. These images are highly compressible, as integrating redundant tokens does not significantly diminish the captured information. Conversely, Figure \ref{fig:model_analysis}(c) highlights images rich in detailed, fine-grained features. Such images demand higher fidelity during visual compression to retain their intricate details.

\section{Discussion}
In this work, we propose DFMR, which dynamically adjusts the compression ratio of visual tokens based solely on the intrinsic content of each image, simultaneously improving both performance and efficiency. Specifically, we introduce a novel metric to quantify image information, determining the optimal number of visual tokens required for effective representation. DFMR achieves improvements in both computational efficiency and model performance. Extensive experiments are conducted on various mainstream multimodal evaluation benchmarks to validate the effectiveness of the proposed method. The method offers a promising solution for extending LLaVA to handle multi-images and video in resource-constrained academic environments and it can also be applied in industry settings for data augmentation to help mitigate the scarcity of open-domain image-text pair datasets in the continued pretraining stage.

During the pretraining stage, token sizes typically reach the trillion level, and the number of GPU cards often scales to hundreds or even thousands within clusters. One of the major computational bottlenecks for GPU utilization in this process is loading and preprocessing the vast datasets before transferring them to the GPUs. To address this, there are two primary approaches: first, pre-transforming and saving the dataset as tokens allow for direct loading onto GPUs, reducing preprocessing time but requiring significant storage for the pre-tokenized data.\cite{shoeybi2019megatron} Second, using dataloaders that support prefetching large amounts of data enables efficient streaming to GPUs, maintaining flexibility without extensive storage needs.\cite{moritz2018ray,leclerc2023ffcv} Our proposed method operates with a computational cost of \( O(1) \), meaning it does not add extra burden to the dataloader and ensures high GPU utilization.

During the continued pretraining stage, maintaining an appropriate dataset mixing ratio is crucial for achieving optimal performance of MLLMs in both domain-specific downstream tasks and general applications\cite{dubey2024llama}. However, open-source image-text pair datasets are relatively scarce compared to domain-specific datasets collected in industry settings, thus becoming a bottleneck for further improving the performance of domain-specific MLLMs. Additionally, generating synthetic image-text pairs is challenging due to the complexity of accurately pairing images with relevant text. The DFMR can be directly applied to data augmentation by setting a manually defined threshold that varies proportionally with the learning rate of the optimization scheduler. For example, using an LLM to generate synthetic text data corresponding to existing images can expand the image-text dataset. Applying DFMR to the expanded dataset by dynamically compressing the same images into different visual lengths with similar textual information provides more tokens for open-domain datasets.

In the LLaVA family, when processing data formats such as multiple images and videos, a certain number of images are sampled, and each image is converted into 576 visual tokens\cite{lin2023video}. If the number of images is large, this can easily exceed the maximum token length predefined by the LLM, especially in resource-constrained academic scenarios. Our proposed method provides a promising solution to this issue by effectively compressing visual tokens, enabling a single model to handle single images, multiple images, and videos without surpassing token length limitations.

\bibliographystyle{unsrt}  
\bibliography{references}  %%% Remove comment to use the external .bib file (using bibtex).
%%% and comment out the ``thebibliography'' section.

\end{document}